\documentclass[letterpaper, 10 pt, conference]{ieeeconf}
\IEEEoverridecommandlockouts
\usepackage{graphics} 
\usepackage{wrapfig} 
\usepackage[linesnumbered,ruled]{algorithm2e}
\graphicspath{{img/}}
\usepackage{epsfig} 
\usepackage{amsmath} 
\usepackage{amssymb}  
\usepackage{amsthm}
\usepackage{bigints}
\usepackage{pifont}
\def\-{\raisebox{.75pt}{-}}
\usepackage{mathtools}
\usepackage{cite}
\usepackage[makeroom]{cancel}
\usepackage[table]{xcolor} 
\usepackage[center]{subfigure}
\usepackage{multirow}
\usepackage{hyperref}
\usepackage{booktabs}
\usepackage{subfigure}
\usepackage{epstopdf}
\usepackage{textcomp}
\usepackage{makecell}
\usepackage{bm}
\usepackage{caption}
\usepackage{algpseudocode,algorithm} 
\usepackage{color, soul}
\usepackage{mathptmx}
\usepackage[11pt]{moresize}
\usepackage{hhline}
\definecolor{ForestGreen}{RGB}{34,139,34}

\newcommand\copyrighttext{%
	\footnotesize This work has been submitted to the IEEE for possible publication. Copyright may be transferred without notice, after which this version may no longer be accessible.}
\newcommand\copyrightnotice{%
	\begin{tikzpicture}[remember picture,overlay]
	\node[anchor=south,yshift=10pt] at (current page.south) {\fbox{\parbox{\dimexpr\textwidth-\fboxsep-\fboxrule\relax}{\copyrighttext}}};
	\end{tikzpicture}%
}
\usepackage{tikz}
\usepackage{textcomp}
\def\BibTeX{{\rm B\kern-.05em{\sc i\kern-.025em b}\kern-.08em
    T\kern-.1667em\lower.7ex\hbox{E}\kern-.125emX}}

\newcommand{\subparagraph}{}
\definecolor{amaranth}{rgb}{0.9, 0.17, 0.31}
\definecolor{bleudefrance}{rgb}{0.19, 0.55, 0.91}

\usepackage{titlesec}
\titlespacing*{\section}{0pt}{1mm plus0.7mm minus0.7mm}{1mm plus0.7mm minus0.7mm}
\titlespacing*{\subsection}{0pt}{1mm plus0.7mm minus0.7mm}{1mm plus0.7mm minus0.7mm}
\titlespacing*{\subsubsection}{0pt}{1mm plus0.7mm minus0.7mm}{1mm plus0.7mm minus0.7mm}
\begin{document}
\title{A Simple and Efficient Multi-task Network for 3D Object Detection and Road Understanding \thanks{$^1$ Mechanical Systems Control Lab, University of California, Berkeley, CA, 94720, USA.}
\thanks{$^2$ Institute of Measurement, Control and Microtechnology, Ulm University, 89081, Ulm, Germany.}
\thanks{Correspondence: \url{di.feng@berkeley.edu}}
}
\author{Di Feng$^{1,2}$, Yiyang Zhou$^{1}$, Chenfeng Xu$^{1}$, Masayoshi Tomizuka$^{1}$, Wei Zhan$^{1}$}
\maketitle

\begin{abstract}
Detecting dynamic objects and predicting static road information such as drivable areas and ground heights are crucial for safe autonomous driving. Previous works studied each perception task separately, and lacked a collective quantitative analysis. In this work, we show that it is possible to perform all perception tasks via a simple and efficient multi-task network. Our proposed network, LidarMTL, takes raw LiDAR point cloud as inputs, and predicts six perception outputs for 3D object detection and road understanding. The network is based on an encoder-decoder architecture with 3D sparse convolution and deconvolution operations. Extensive experiments verify the proposed method with competitive accuracies compared to state-of-the-art object detectors and other task-specific networks. LidarMTL is also leveraged for online localization. Code and pre-trained model have been made available at \url{https://github.com/frankfengdi/LidarMTL}.
\end{abstract}

\copyrightnotice

\section{Introduction}\label{sec:introduction}

Reliable traffic object detection and road understanding near the ego-vehicle are fundamental perception problems in autonomous driving. Movable objects are often perceived at the instance level with class labels and bounding boxes, or at the 2D image pixel or 3D point level depending on sensing modalities. A comprehensive road understanding requires the perception algorithm to identify drivable areas, lane markings, and road shapes, to name a few. Furthermore, all these perception tasks need to run accurately and quickly for online deployment. However, most existing methods, especially those using deep learning approaches, focus on improving each task separately, with task-specific network architectures and evaluation metrics. This task-specific solution is inefficient when dealing with multiple tasks. While high inference speed might maintain via parallel computing, the memory footprints and computation costs scale linearly with the number of networks, which quickly become infeasible with limited hardware resources. 

Multi-Task Learning (MTL) provides a strategy to largely reduce memory footprints and computation costs by performing all tasks via a unified model in one forward pass~\cite{vandenhende2020multi}. In deep learning, MTL translates to learn the shared representation of multiple tasks typically via an encoder-decoder network architecture. 
MTL was applied to 2D object detection and road understanding using RGB camera images~\cite{teichmann2018multinet,qian2019dlt,chen2018driving} and has been recently introduced to 3D perception using Lidar point clouds~\cite{yang2018hdnet,liang2019multi,yan2020lidar}. 
\begin{figure}[!tpb]
	\centering
	\includegraphics[width=8cm]{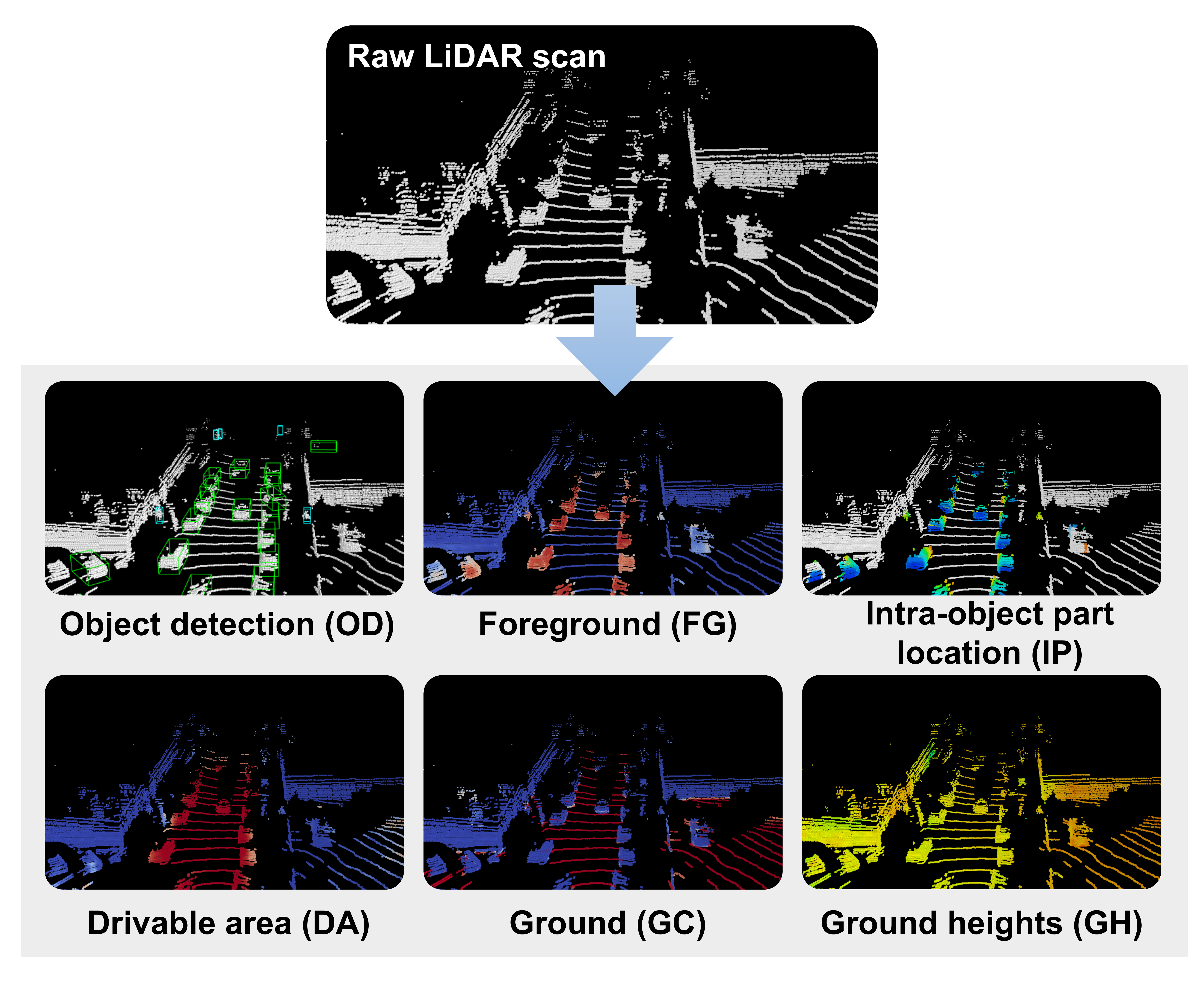}
	\caption{The proposed multi-task network, LidarMTL, takes raw Lidar point clouds as inputs, and performs six 3D perception tasks in one forward pass in a frame-by-frame manner. The tasks include object detection (OD), foreground point classification (FG), intra-object part location regression (IP), object-free drivable area classification (DA), ground area classification (GC), ground height estimation (GH).} \label{fig:introduction}
	\vspace{-5mm}
\end{figure}

In this work, we propose a Lidar-based multi-task learning network called LidarMTL to jointly perform six perception tasks for 3D object detection and road understanding, as shown by Fig.~\ref{fig:introduction}. Objects are detected with class labels and 3D bounding boxes (task OD). Furthermore, their associated Lidar points are segmented as foreground (task FG), and their relative locations to object centroids are regressed (task IP). Road perception includes point-wise drivable area and ground area classification (tasks DA and GC) as well as ground height estimation (task GH). 

The benefits of those perception tasks have been studied in previous works. For example, FG and IP are leveraged to refine bounding boxes in the second-stage of a two-stage object detector~\cite{shi2020part}, as they provide useful point-level semantic and geometrical information of objects. GC and GH are used to remove ground~\cite{li2020integrate} or normalize the heights of Lidar points~\cite{liang2019multi}, especially when the ground is not flat. DA provides strong road priors to reduce false-positive predictions in object detection~\cite{yang2018hdnet} and motion forcasting~\cite{chang2019argoverse}. Unlike previous works which explore each perception task separately, we show that it is possible to perform all perception tasks efficiently and accurately via a unified network. Besides, previous works such as~\cite{shi2020part,liang2019multi} focus on how to employ one or several perception tasks as auxiliary tasks to support the target task, without analyzing the performance of those auxiliary tasks. In this work, we consider each perception task of equal importance, and conduct comprehensive experiments to analyze their performance in the single-task and multi-task settings. 

In principle, the LidarMTL network works by adding task-specific heads to a 3D UNet architecture and training the full network with a multi-task loss in an end-to-end manner. UNet is a well-performed encoder-decoder network widely applied to 2D image segmentation. Following~\cite{shi2020part}, we extend UNet to efficiently process 3D Lidar points represented as voxels with 3D sparse convolution and deconvolution operations. The resulting network has only $6.5$M trainable parameters and runs at an inference speed of $30$FPS on a Titan RTX GPU, which is $2\times$ smaller and $6\times$ faster than performing all tasks sequentially using task-specific networks. Extensive experiments on the Argoverse Dataset~\cite{chang2019argoverse} shows that the LidarMTL network achieves competitive accuracies compared to state-of-the-art object detectors and other task-specific networks. The network is also employed to substantially improve online localization. 


\section{Related Works}
\label{sec:related_works}
\subsection{Lidar-based Object Detection}
Lidar point cloud is usually represented by 2D projected images~\cite{yang2018pixor,chen2017multi}, raw Lidar point~\cite{shi2019pointrcnn,xu2017pointfusion}, and voxels~\cite{zhou2017voxelnet}. Compared to the other methods, voxel representation can not only be processed efficiently using 3D sparse convolution~\cite{yan2018second}, but also preserve approximately similar information to raw point cloud with small voxel size. Therefore, voxel-based backbone networks have been widely applied to learn Lidar features in conjunction with 2D CNN detection head~\cite{yan2018second,shi2020part,shi2020pv,he2020structure}. A special case is PointPillars~\cite{lang2018pointpillars}, which efficiently processes Lidar points by vertical 3D columns called pillars. Our proposed LidarMTL network follows this ``voxel-based backbone + 2D CNN detection head'' pipeline to perform object detection.

\subsection{Road Understanding}
Understanding the 3D road information online is crucial for safe autonomous driving, especially when HD maps are not available. A variety of methods have been proposed for online mapping, such as road area classification~\cite{caltagirone2017fast,fan2020sne}, lane and boundary detection~\cite{bai2018deep,liang2019convolutional}, ground plane estimation~\cite{chen20153d}, road topology recognition~\cite{baumann2018classifying,oeljeklaus2017combined}, and road scene semantic segmentation~\cite{milioto2019rangenet++,xu2020squeezesegv3,roddick2020predicting}. In~\cite{yan2020lidar}, a multi-task network is designed for multiple object-free road perception tasks, including drivable area classification, road height estimation, and road topology classification.

\subsection{Joint Object Detection and Road Understanding}
Existing methods usually follow the hard parameter sharing scheme~\cite{vandenhende2020multi}, where networks consist of a shared encoder and several task-specific decoders. MultiNet~\cite{teichmann2018multinet} jointly performs object detection, street recognition, and road area classification. It is built by a large 2D CNN encoder based on the VGG16 or ResNet backbones, followed by task-specific branches with several convolution layers. DLT-Net~\cite{qian2019dlt} follows the similar architecture for object detection, road area classification, and lane detection. Besides, HDNet~\cite{yang2018hdnet} and MMF~\cite{liang2019multi} propose to use drivable road maps or ground heights as auxiliary inputs for Lidar-based object detectors, in order to improve the detection accuracy by adding road priors. Our proposed LidarMTL also uses the hard parameter sharing: a detection head and a decoder with sparse deconvolutions are added to the encoder for object detection and point-wise predictions, respectively.

\section{Methodology} \label{sec:methodology}
\begin{figure*}[!tpb]
	\centering
	\includegraphics[width=0.92\linewidth]{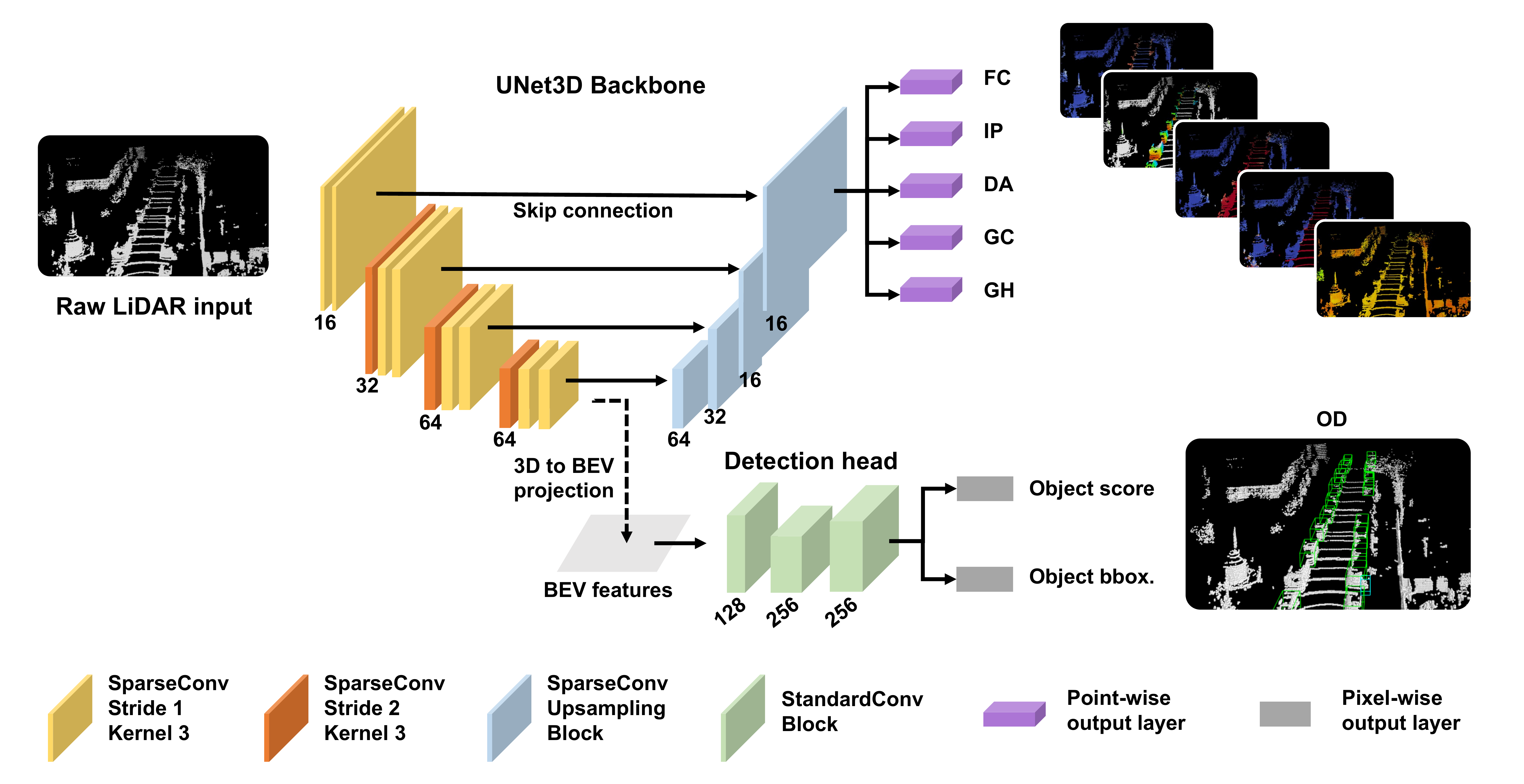}
	\caption{The LidarMTL network architecture. The network jointly performs object detection (OD) and five point-wise perception tasks, namely, foreground classficiation (FC), Intra-object part location regression (IP), drivable area classification (DA), ground classification (GC), and ground height estimation (GH). The network is based on a UNet backbone with 3D sparse convolution and deconvolution. A small detection head with 2D convolution is added to perform object detection on the Lidar Bird's Eye View (BEV).} \label{fig:network}
	\vspace{-3mm}
\end{figure*}
\subsection{Task Definition}
The proposed Lidar-based multi-task learning network, LidarMTL, jointly performs six perception tasks via a single feed-forward pass, namely, 3D object detection (OD), foreground classification (FG), intra-object part location regression (IP), drivable area classification (DA), ground area classification (GC), and ground height regression (GH). The method is developed based on the Argoverse dataset~\cite{chang2019argoverse}, because to our knowledge it is the only public dataset that provides both dynamic object labels and static map information with ground heights and drivable areas. 

More specifically for these tasks, Lidar points within the bounding boxes of the dynamic objects are regarded as foreground. Intra-part object locations are defined as the positions of \textit{foreground} points relative to their corresponding object's centroids. Driveable areas are \textit{object-free} regions which could be driven by vehicles. Ground height estimation is performed both for ground areas and non-ground areas (such as foreground and buildings). 

\subsection{Overview}
We aim to design a simple and efficient multi-task network for joint 3D object detection and road understanding. In this regard, the LidarMTL is based on the voxelized Lidar point cloud representation and the UNet backbone with 3D sparse convolution and deconvolutions (we name the model ``UNet3D''). Fig.~\ref{fig:network} shows the network architecture. The 3D space is voxelized into regular voxels, with no-empty voxels being encoded with Lidar features. The voxelized Lidar point cloud is processed by the UNet backbone network with the encoder-decoder architecture~\cite{ronneberger2015u}. The encoder consists of several 3D sparse convolutions, and downsamples the input spatial resolution by $8$ times in order to extract high-level Lidar features. The decoder gradually upsamples the Lidar features to the original spatial resolution via 3D sparse deconvolutions. We choose the UNet backbone network and voxelized Lidar representation following the idea from~\cite{shi2020part}. The network well-preserves the geometric information of Lidar points by setting a proper voxel size, and has been shown in~\cite{shi2020part} to achieve higher efficiency than the raw point-based methods (such as PointRCNN~\cite{shi2019pointrcnn}).

Point-wise predictions are made by adding output layers directly on the decoder network, including tasks FG, IP, DA, GC, and GH. To perform object detection (OD), the 3D Lidar features from the encoder are projected onto the Bird's Eye View (BEV), and then processed by a detection head with several standard 2D convolution layers for classification score prediction and bounding box regression. Note that employing 2D CNN on Lidar BEV features is a common way to do object detection~\cite{yan2018second,lang2018pointpillars,shi2020pv}. Besides, it is found more effective than performing object detection from the decoder network (cf. Sec~\ref{experimental_result:object_detection}). 
\subsection{Input and Output Representation}
Given an Lidar scan, let $y$ be the target output. The input features of each voxel are encoded as the mean values of the Lidar point positions in the Lidar coordinate system. The perception tasks FC, DA, and GC are formulated as the point-wise binary classification problems, with their labels $y_{\text{FG}}, y_{\text{DA}}, y_{\text{GC}}=1$ indicating positive samples, and $0$ negative samples. The tasks IP and GH are formulated as the point-wise regression problems, with $y_{\text{IP}}=[x',y',z']$ a continuous vector indicating 3D Lidar point locations relative to their corresponding object centroids, and $y_{\text{GH}}$ the ground heights. As for OD, the label $y_{\text{OD}}$ consists of object classes $y_{\text{cls}}$, and bounding box regression variables $y_{\text{bbox}}$, i.e. $y_{\text{OD}}= [y_{\text{cls}}, y_{\text{bbox}}]$. Bounding boxes are parameterized by $y_{\text{bbox}}=[\Delta x, \Delta y, \Delta z, \Delta l, \Delta w, \Delta h, \Delta \theta]$, with $\Delta x, \Delta y, \Delta z$ being the residual centroid 3D positions, $\Delta l, \Delta w, \Delta h$ the residual length, width, and height, and $\Delta \theta$ the residual orientation relative to pre-defined anchors. The network makes predictions for $y_{\text{FG}}, y_{\text{DA}}, y_{\text{GC}}, y_{\text{cls}}$ via the softmax function, and directly regresses the bounding box parameters. Following~\cite{shi2020part}, $y_{\text{IP}}$ are normalized to be within $[0,1]^3$ and are predicted by the softmax function as well, as this encoding strategy is found more stable than direct regression. 
\subsection{UNet3D Backbone}
As shown in Fig.~\ref{fig:network}, the encoder in the backbone processes the voxelized Lidar point cloud by four stages of 3D sparse convolutions with increasing feature dimensions $16,32,64,64$. The network downsamples the spatial resolution by $8$ times through three sparse convolution layers~\cite{yan2018second} with stride $2$, each followed by two submanifold sparse convolution layers~\cite{graham2017submanifold} with stride 1. The decoder consists of four upsampling blocks with decreasing feature dimensions $64,32,16,16$ and strides $2,2,2,1$. In each block, the features from the previous block are combined with the skip-connected features from the encoder via concatenation, and are further processed by a submanifold sparse convolution layer and a sparse inverse convolution layer, in order to upsample the spatial resolution. All convolution and deconvolution layers in the backbone have a kernel size of $3\times3\times3$. Finally, task-specific $1\times1\times1$ sparse convolution layers are added to the last upsampling block for point-wise predictions.
\subsection{Detection Head}
The detection head projects the 3D Lidar features from the UNet3D encoder to the Bird's Eye View (BEV), and processes the BEV features through three 2D convolution blocks. The first block consists of six standard convolution layers with feature dimension $128$ and stride $1$. The second block increases feature dimension to $256$. It downsamples the spatial resolution by a convolution layer with stride $2$, stacked with five 2D convolution layers with stride $1$. The last block is an upsampling layer with dimension $256$ and stride $2$. All convolution layers in the detection head have a kernel size of $3\times3$. The classification scores and bounding boxes are predicted by the output layers with $1\times1$ convolution. 

Similar to~\cite{yan2018second}, the object detector regresses residual bounding box parameters relative to the pre-defined 3D anchors with fixed sizes, because objects from the same category are of approximately similar sizes. For each pixel and object category, we place two anchors with rotations of $0$ and $90$ degrees, with their sizes being the mean values from all ground truths in the Argoverse dataset. 

\subsection{Joint Training}
The full network is trained end-to-end via a multi-task loss function. Denote $L$ as a loss function. For an input data frame, the multi-task loss function, $L_{\text{MTL}}$, is formulated as a weighted sum of the task-specific losses:
\begin{equation}\label{eq:standard_mtl_loss}
    L_{\text{MTL}} = \sum_{\substack{i \in\{\text{OD,FG,IP,}\\  \text{\ \ \ \ \ \ \ \ \ DA,FC,GH}\}}} w_i L_i,
\end{equation}
where $w_i$ and $L_i$ represent the task-specific loss weights and loss functions, respectively. To learn $y_{\text{DA}}, y_{\text{GC}}, y_{\text{IP}}$, we use the standard cross entropy loss. As for $y_{\text{FG}}$ and $y_{\text{cls}}$, we use the focal loss~\cite{lin2018focal} due to the large positive-negative sample imbalance problem. Finally, $y_{\text{GH}}$ and $y_{\text{bbox}}$ are learnt by the standard $L_1$ loss.

A loss weights $w_i$ controls the influence of a task. It can be pre-defined through grid search, or optimized by task balancing approaches~\cite{vandenhende2020multi}. In this work, we employ the uncertainty weighting strategy proposed by Kendall \textit{et al.}~\cite{cipolla2018multi}. It uses the task-dependent uncertainty, parameterized by the noise parameter $\sigma$, to balance the single-task losses. Such noise parameters are jointly optimized during training, resulting in an adaptive multi-task loss function $L^{\text{adaptive}}_{\text{MTL}}$ written as:
\begin{equation}\label{eq:uncertainty_aware_mtl_loss}
    L^{\text{adaptive}}_{\text{MTL}} = \sum_{\substack{i \in\{\text{OD,FG,IP,}\\  \text{\ \ \ \ \ \ \ \ \ DA,FC,GH}\}}} \frac{1}{2\sigma^2_i} L_i + \frac{1}{2}\log \sigma^2_i \ .
\end{equation}

\section{Experimental Results} \label{sec:experimental_result}
The experimental results are structured as follows. In Sec.~\ref{experimental_result:performance_evaluation}, we evaluate the performance of each perception task separately. We compare the proposed multi-task network with single-task networks, and show its benefits in achieving on-par performance with task-specific networks, but with substantially lower memory footprints and higher inference speed. Afterwards, we conduct ablation studies in Sec.~\ref{experimental_result:ablation_study} regarding the number of tasks and the loss weights, and test the network's robustness with downsampled Lidar points. Finally, we demonstrate in Sec.~\ref{experimental_result:application_online_localization} that our proposed multi-task network provides useful semantics which largely improve online localization.  

\subsection{Experimental Setup}
\subsubsection{Dataset}
All experiments are conducted on the Argoverse 3D Tracking Dataset~\cite{chang2019argoverse}, which was recorded in Miami and Pittsburgh in the USA under various weather conditions and times of a day. The dataset provides 3D bounding boxes and tracks annotations, with RGB images from seven cameras, Lidar point clouds from two 32-beam Velodyne Lidar sensors, as well as HD maps annotating drivable areas, ground heights, ground areas and center-lines. 
For the object detection task, we focus on the ``VEHICLE'' and ``PEDESTRAIN'' classes. For the point-wise perception tasks, we prepare the ground truth labels for each Lidar point. The data was recorded in sequence with length varying from $15$ to $30$ seconds ($10$ Hz). To reduce the sequential dependency between frames, we down-sample the dataset by a factor of $5$. 
The resulting dataset we use contains $2609$ training frames and $996$ evaluation frames, with over $20$K VEHICLE and $6.7$K PEDESTRIAN objects. 

\subsubsection{Implementation Details}
All networks are trained with the same optimization settings from scratch up to $80$ epochs. The ADAM optimizer is used with an initial learning rate of $0.01$, a step decay of $0.1$, and a batch size of $4$. 
In order to have a fair comparison with state-of-the-art object detectors (such as PV-RCNN~\cite{shi2020pv} and PointPillars~\cite{lang2018pointpillars}), which only process Lidar point clouds on the camera front-view, we extract Lidar point clouds corresponding to synchronized front-view images from the original Argoverse dataset, and train the front-view networks for most experiments. In this regard, we use the Lidar point cloud within the range length $\times$ width $\times$ height = $[0,70.4]$m$\times[-40,40]$m$\times[-1.5,4,0]$m, and do discretization at $0.1$ meter voxel resolution. Besides, to employ our proposed LidarMTL network in online localization, we train a full-range network that processes Lidar point cloud within the range $[-70.4,70.4]$m$\times[-70.4,70.4]$m$\times[-1.5,4,0]$m. All experiments are conducted using a Titan RTX GPU. The inference time for the front-view LidarMTL reaches $30$FPS and for the full-range LidarMTL $7.7$FPS.
\subsection{Performance Evaluation}\label{experimental_result:performance_evaluation}
\subsubsection{Object Detection (\textbf{OD})}\label{experimental_result:object_detection}
\begin{table*}[tbp]
	\centering
	\resizebox{1.00\linewidth}{!}{\begin{tabular}{l|ll|cc|cc|cc}
			\Xhline{2\arrayrulewidth}
			\multirow{2}{*}{Methods} & \multicolumn{2}{c|}{VEHICLE} & \multicolumn{2}{c|}{PEDESTRAIN} & \multirow{2}{*}{$\text{mAP}_{BEV}(\%)$} & \multirow{2}{*}{$\text{mAP}_{3D}(\%)$} & Trainable & Inference \\ \cline{2-5}
			& $\text{AP}_{BEV}@0.7(\%)$ & $\text{AP}_{3D}@0.7(\%)$ & $\text{AP}_{BEV}@0.5(\%)$ & $\text{AP}_{3D}@0.5(\%)$ & & & param. (M) & speed (FPS) \\ \Xhline{2\arrayrulewidth}  
			PV-RCNN~\cite{shi2020pv} & $77.5,62.0, 21.1$ & $63.2,38.0,3.8$ & $51.8,26.6,4.5$ & $45.7,22.2,3.0$ & $56.1$ & $43.2$ & $13.10$ & $14.6$ \\
			PointPillars~\cite{lang2018pointpillars} & $75.3, 57.2, 16.6$ & $53.5, 27.8, 2.7$ & $37.4, 22.3, 4.0$ & $30.3, 16.8, 2.3$ & $51.5$ & $35.3$ & $4.82$ & $71.5$ \\
			Second~\cite{yan2018second} & $72.0, 53.9, 14.1$ & $50.9, 25.0, 1.9$ & $41.1, 22.8, 5.0$ & $33.6, 17.5, 2.6$ & $49.7$ & $35.1$ & $5.31$ & $54.2$ \\
			UNet3D & $73.0, 35.9, 4.4$ & $50.9, 13.7, 0.5$ & $56.7, 25.1, 2.9$& $ 44.4, 17.4, 1.5$ & $46.4$ & $32.4$ & $1.90$ & $33.2$ \\
			LidarBEV & $71.8, 56.1, 14.0$ & $50.3, 23.8, 1.7$ & $42.0, 22.9, 4.4$& $36.6, 16.3, 2.1$ & $49.9$ & $34.6$ & $5.31$ & $59.6$ \\
			\rowcolor{lightgray!15} LidarMTL & $72.9, 56.9, 14.1$ & $53.4, 24.3, 1.8$ & $40.6, 22.9, 6.1$ & $33.3, 17.0, 4.2$ & $49.8$ & $35.0$ & $6.52$ & $30.0$ \\
			\Xhline{2\arrayrulewidth} \end{tabular}}
	\caption{A comparison of Object Detection (OD) performance, as well as the number of trainable parameters and inference speed. Detections are grouped into different Lidar ranges ($0-30$m, $30-50$m, $50-70$m). The AP scores are measured at IOU=0.7 threshold for ``VEHICLE'' class, and IOU=0.5 for ``PEDESTRAIN'' class.} \label{tab:detection_results}
	\vspace{-5mm}
\end{table*}
We evaluate the object detection performance using the standard Average Precision for 3D detection ($\text{AP}_{3D})$ and on the Bird's Eye View (BEV) ($\text{AP}_{BEV}$). They are measured at the Intersection Over Union IOU=0.7 threshold for ``VEHICLE'' objects and IOU=0.5 for ``PEDESTRAIN'' objects, respectively, as suggested by~\cite{Geiger2012CVPR}. The IOU scores in object detection are geometric overlap ratios between bounding boxes, and indicate the localization accuracy. We report the AP scores with respect to increasing Lidar ranges ($0-30$m, $30-50$m, and $50-70$m), as well as the mean AP scores, mAP, by averaging over all distances and object classes (similar to~\cite{everingham2010pascal}). Besides, we report the number of trainable parameters and the inference speed for each object detectors. Tab.~\ref{tab:detection_results} summarizes the results.

The proposed multi-task network (LidarMTL) is compared against several Lidar-based object detectors. The LidarBEV network follows the same detection architecture with LidarMTL (Encoder with sparse 3D convolution + BEV detection head with 2D convolution). It serves as the baseline to study the object detection performance when introducing multiple tasks. The UNet3D network directly employs the encoder-decoder architecture from LidarMTL to predict object classes and bounding boxes on each Lidar point (without BEV detection head and pre-defined anchors), and is used to verify the network architecture design. Furthermore, we re-train state-of-the-art detectors, PV-RCNN~\cite{shi2020pv}, PointPillars~\cite{lang2018pointpillars}, and Second~\cite{yan2018second}, using our experimental setup. Note that PV-RCNN is a two-stage object detector, whereas all other detectors are one-stage. UNet3D directly regresses bounding box parameters, whereas the others are based on pre-defined anchors and BEV detection heads.

As Tab.~\ref{tab:detection_results} illustrates, the proposed LidarMTL network achieves similar detection accuracy to LidarBEV, SECOND, and PointPillars, with comparable number of parameters ($6.52$M) and reasonable inference speed ($30$FPS). PV-RCNN has the highest AP scores compromised by over $2\times$ more parameters and computation cost compared to LidarMTL. Though UNet3D has only $1.9$M parameters, it has the worst detection accuracy with $2-3\%$ smaller $mAP$ scores compared to LidarMTL, indicating the importance of adding anchor priors and BEV detection heads for precise object detection. In conclusion, the proposed LidarMTL shows competitive detection performance to other detectors regarding accuracy, model size, and inference speed. 

\subsubsection{Foreground (\textbf{FG}), Drivable Area (\textbf{DA}), and Ground Classification (\textbf{GC})}
We evaluate the foreground, drivable area, and ground classification tasks using the Average Precision (AP), Intersection Over Union (IOU), and classification accuracy scores at $0.5$ probability threshold. Those evaluation metrics measure the classification performance at each Lidar point, and have been used as the standard metrics for road detection~\cite{Fritsch2013ITSC} or semantic segmentation~\cite{everingham2010pascal}. Note that unlike the IOU metric for object detection in the previous section, here an IOU score is measured by $\text{IOU} = 100*\text{TP}/(\text{TP+FP+FN})$ according to~\cite{everingham2010pascal}, with TP, FP, FN being the number of points categorized as true positive, false positive, and false negative samples. For each task, a task-specific UNet3D network is trained to compare with the LidarMTL network. Furthermore, since these point-wise perception tasks can be regarded as semantic segmentation, we additionally train two state-of-the-art semantic segmentation networks, namely, RangeNet++~\cite{milioto2019rangenet++} and SqueezeSegv3~\cite{xu2020squeezesegv3}, to perform foreground and ground classification. We do not conduct experiments for drivable area classification, because it is not mutually exclusive with other two tasks. 

Experimental results are shown in Tab.~\ref{tab:foreground_classification_results}, Tab.~\ref{tab:drivable_area_classification_results}, and Tab.~\ref{tab:ground_classification_results}. The proposed multi-task network achieves on-par performance with the single-task network, with less than $1\%$ difference in all evaluation criterion. The network also shows competitive results with RangeNet++ and SqueezeSegv3, verifying the effectiveness of the network architecture design. 

\subsubsection{Ground Height Estimation (\textbf{GH})}\label{experimental_result:ground_height_estimation}
We evaluate the ground height estimation performance using the Root Mean Squared Errors (RMSE) and Mean Average Errors (MAE) metrics, which are widely used in the depth prediction task for RGB camera images~\cite{Uhrig2017THREEDV}. Tab.~\ref{tab:ground_heights_all_results} reports the performance for all Lidar points, grouped with respect to the Lidar ranges ($0-30$m, $30-50$m, $50-70$m). The LidarMTL network is compared with the task-specific UNet3D network, as well as a simple heuristic which assumes a ground plane given the ego-vehicle's pose information. Note that the ground plane assumption has been widely used to remove ground Lidar points for object detection~\cite{ku2017joint,chen20153d}. 

Tab.~\ref{tab:ground_heights_all_results} shows that the ground plane method results in larger errors at longer distances, indicating that the ground is not flat. The errors produced from the UNet3D and LidarMTL networks are much smaller than the ground plane method (over $40\%$ RMSE reduction for all points), showing the benefits of the point-wise ground height estimation. The LidarMTL network produces slightly larger errors than the UNet3D network ($<2$cm), showing small negative transfer phenomena often seen in multi-task learning~\cite{vandenhende2020multi}.

A more interesting experiment is to evaluate the ground height estimation for Lidar points which belong to objects. Such information could be used to normalize objects' heights and has the potential to improve detection performance, as shown in~\cite{liang2019multi}. Tab.~\ref{tab:ground_heights_objects_results} shows that both networks predict ground heights accurately, with RMSE errors smaller than $20$cm even at $50-70$m range. 

\subsubsection{Intra-object part locations (\textbf{IP})}
Finally, we evaluate the performance for intra-object part location predictions, with the same evaluation metrics (RMSE and MAE) used in the previous section (Sec.~\ref{experimental_result:ground_height_estimation}). Both LidarMTL and UNet3D networks perform similarly, with the LidarMTL network producing slightly smaller errors ($<1$) at long distance ($50-70$m) than the UNet3D network.

\begin{table*}
\centering
\setlength\tabcolsep{4.5pt}
\parbox[t]{0.32\linewidth}{
\centering
\begin{tabular}[t]{l|c c c}
\Xhline{2\arrayrulewidth}
Methods& $AP (\%)$ & IOU $(\%)$& Accu. $(\%)$ \\ \Xhline{2\arrayrulewidth}  
RangeNet++~\cite{milioto2019rangenet++} & - & $82.4$ & - \\
SqueezeSegv3~\cite{xu2020squeezesegv3}& - & $84.2$ & - \\
UNet3D& $96.2$ & $85.4$ & $98.7$ \\
\rowcolor{lightgray!15} LidarMTL & $97.0$ & $85.6$ & $98.7$ \\
\Xhline{2\arrayrulewidth} \end{tabular}
\caption{Foreground (FG).} \label{tab:foreground_classification_results}
}
\hfill
\parbox[t]{0.65\linewidth}{
\centering
\begin{tabular}[t]{l|c c c c| c c c c}
\Xhline{2\arrayrulewidth}
\multirow{2}{*}{Methods} & \multicolumn{4}{c|}{RMSE (cm)}  & \multicolumn{4}{c}{MAE (cm)} \\ \cline{2-9}
& All & 0-30m & 30-50m & 50-70m & All & 0-30m & 30-50m & 50-70m \\ \Xhline{2\arrayrulewidth}
Plane & $31.2$ & $21.0$ & $38.0$ & $53.5$ & $21.6$ & $16.3$ & $28.2$ & $35.5$ \\
UNet3D& $17.8$ & $7.9$ & $21.0$ & $40.0$ & $7.8$ & $4.9$ & $9.8$ & $20.1$ \\
\rowcolor{lightgray!15} LidarMTL & $18.6$ & $8.8$ & $22.2$ & $40.4$ & $8.8$ & $5.7$ & $11.0$ & $21.4$  \\
\Xhline{2\arrayrulewidth} \end{tabular}
\caption{Ground heights (GH) (all Lidar points).} \label{tab:ground_heights_all_results}
}
\vfill
\parbox[t]{0.32\linewidth}{
\centering
\begin{tabular}[t]{l|c c c}
\Xhline{2\arrayrulewidth}
Methods & $AP (\%)$& IOU (\%)& Accu. $(\%)$ \\ \Xhline{2\arrayrulewidth}
RangeNet++~\cite{milioto2019rangenet++} & - & $95.2$ & - \\
SqueezeSegv3~\cite{xu2020squeezesegv3} & - & $95.9$ & - \\
UNet3D& $99.6$ & $94.5$ & $98.2$ \\
\rowcolor{lightgray!15} LidarMTL & $99.6$ & $94.0$ & $98.0$ \\
\Xhline{2\arrayrulewidth} \end{tabular}
\caption{Ground areas (GC).} \label{tab:ground_classification_results}
}
\hfill 
\parbox[t]{0.65\linewidth}{
\centering
\begin{tabular}[t]{l|c c c c| c c c c}
\Xhline{2\arrayrulewidth}
\multirow{2}{*}{Methods} & \multicolumn{4}{c|}{RMSE (cm)}  & \multicolumn{4}{c}{MAE (cm)} \\ \cline{2-9}
& All & 0-30m & 30-50m & 50-70m & All & 0-30m & 30-50m & 50-70m \\ \Xhline{2\arrayrulewidth}
Plane & $20.8$ & $17.8$ & $27.3$ & $35.4$ & $15.3$ & $12.9$ & $22.4$ & $28.7$ \\
UNet3D & $8.6$ & $6.5$ & $11.7$ & $19.1$ & $5.5$ & $4.4$ & $8.0$ & $13.9$ \\
\rowcolor{lightgray!15} LidarMTL & $9.8$ & $8.2$ & $12.2$ & $19.6$ & $6.7$ & $5.8$ & $8.9$ & $14.7$  \\
\Xhline{2\arrayrulewidth} \end{tabular}
\caption{Ground heights (GH) (only foreground points).} \label{tab:ground_heights_objects_results}
}
\vfill
\parbox[m]{0.32\linewidth}{
\centering
\begin{tabular}[t]{l|c c c}
\Xhline{2\arrayrulewidth}
Methods & $AP (\%)$& IOU (\%)& Accu. $(\%)$ \\ \Xhline{2\arrayrulewidth}
UNet3D& $97.9$ & $86.5$ & $94.2$ \\
\rowcolor{lightgray!25} LidarMTL & $97.4$ & $84.5$ & $93.4$ \\
\Xhline{2\arrayrulewidth} \end{tabular}
\caption{Drivable areas (DA).} \label{tab:drivable_area_classification_results}
}
\hfill
\parbox[m]{0.65\linewidth}{
\centering
\begin{tabular}[t]{l|c c c c| c c c c}
\Xhline{2\arrayrulewidth}
\multirow{2}{*}{Methods} & \multicolumn{4}{c|}{RMSE}  & \multicolumn{4}{c}{MAE} \\ \cline{2-9}
& All & 0-30m & 30-50m & 50-70m & All & 0-30m & 30-50m & 50-70m \\ \Xhline{2\arrayrulewidth}  
UNet3D& $10.0$ & $8.1$ & $13.5$ & $18.8$ & $5.6$ & $4.6$ & $8.0$ & $13.8$ \\
\rowcolor{lightgray!15} LidarMTL & $9.9$ & $8.2$ & $13.3$ & $18.1$ & $5.7$ & $4.7$ & $8.0$ & $13.2$  \\
\Xhline{2\arrayrulewidth} \end{tabular}
\caption{Intra-object part locations (IP).} \label{tab:intra_part_locations_results}
}
\vspace{-5mm}
\end{table*}
\begin{figure}[!htpb]
    \centering
    \begin{minipage}{1\linewidth}
	\centering
	\subfigure[]{\label{fig:model_size_comparison}\includegraphics[width=0.48\textwidth]{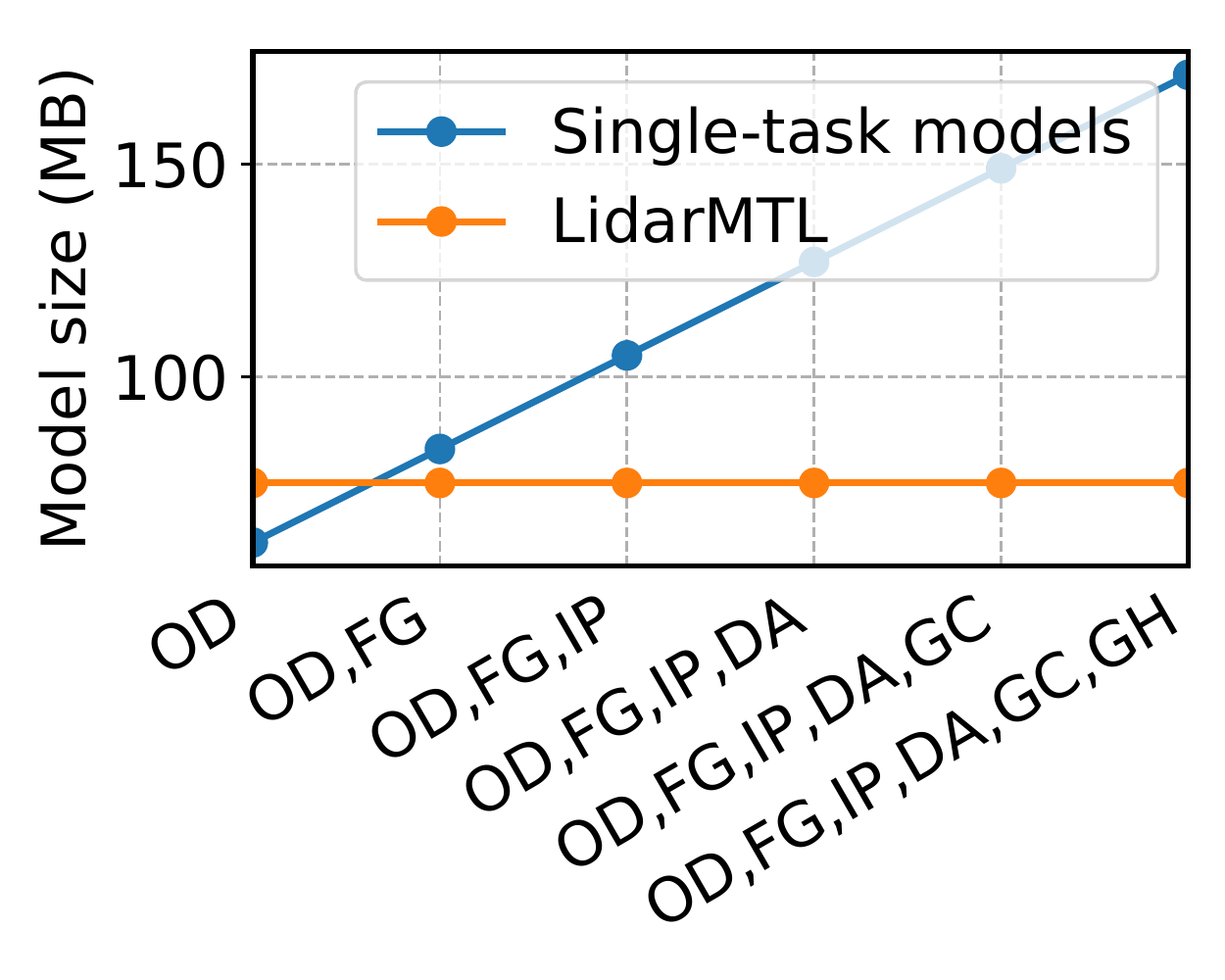}}
    \subfigure[]{\label{fig:efficiency_comparison}\includegraphics[width=0.48\textwidth]{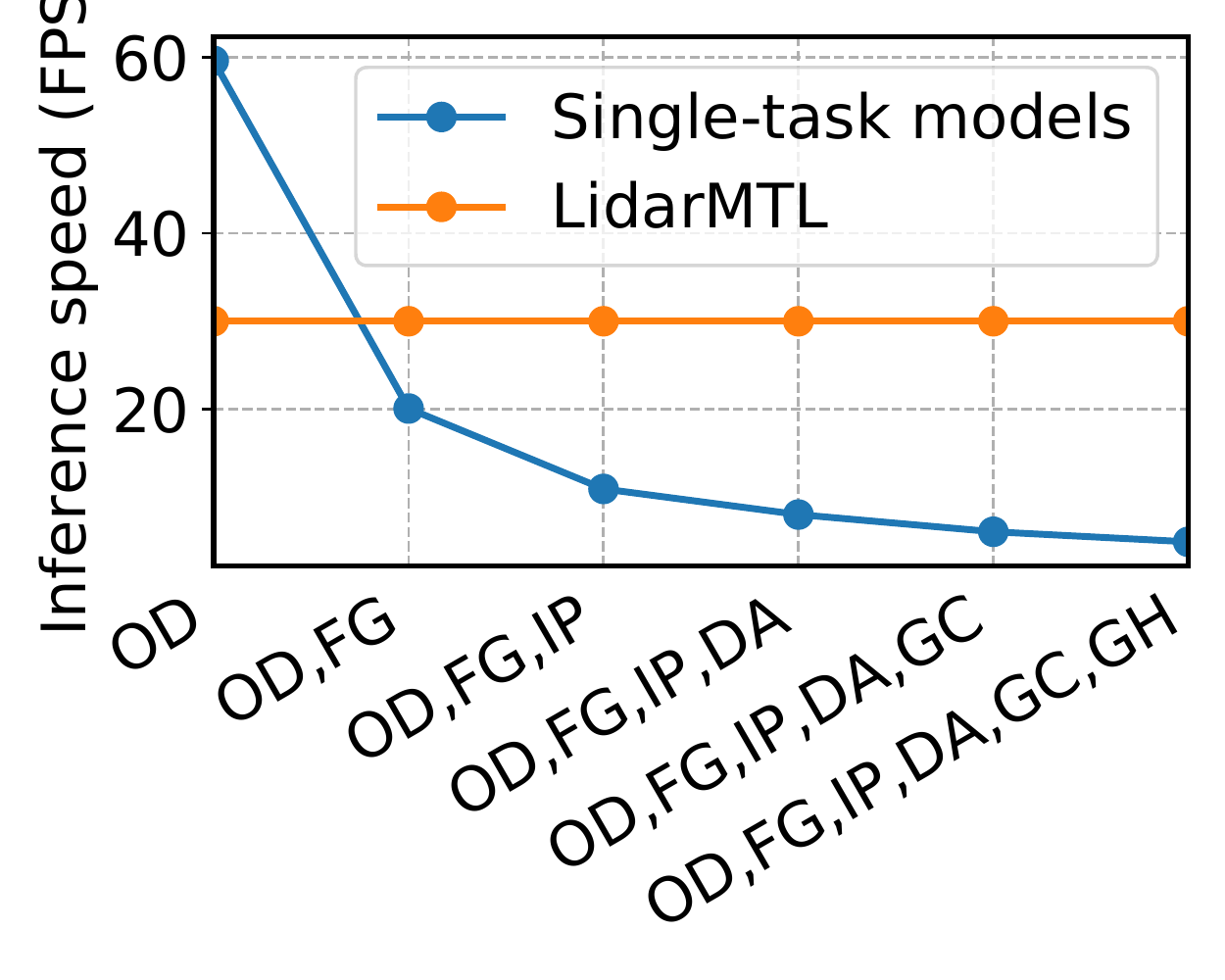}}
	\caption{A comparison of model size (in MegaByte) and inference speed (in FPS) required to achieve multiple perception tasks, by employing the proposed multi-task network or the ``Single-task models'' which performs all tasks separately by a chain of task-specific networks.}
\end{minipage}
\vspace{-3mm}
\end{figure}
\subsubsection{Model Size and Inference Speed}
We quantitatively show the benefits of lower memory footprints and higher inference speed brought by the proposed multi-task network, compared to the  ``Single-task models'', which performs all tasks separately by a chain of task-specific networks. In this regard, we employ the LidarBEV network introduced in~\ref{experimental_result:object_detection} for object detection, and train UNet3D networks for other tasks. Starting from object detection, we gradually increase the number of perception tasks, and calculate the required memory footprints and the inference speed averaged over all predictions on the evaluation data. Fig.~\ref{fig:model_size_comparison} and Fig.~\ref{fig:efficiency_comparison} show the model size (in MegaByte) and the inference speed (in FPS), respectively. The LidarMTL network outperforms the single-task models approach, when considering more than one perception task. While the LidarMTL network remains constant model size and inference speed regardless of the number of tasks, the single-task models approach requires linearly-increasing memory and much lower inference speed. When performing all six perception tasks, the multi-task network is more than $2\times$ smaller and $6\times$ faster, showing its high efficiency, which is critical for online deployment.
\subsection{Ablation Study}\label{experimental_result:ablation_study}
\subsubsection{Number of Tasks}
\begin{figure}[tpb]
    \centering
    \begin{minipage}{1\linewidth}
	\centering
	\subfigure[OD + Multi-tasks]{\label{fig:detection_mtl_ablation}\includegraphics[width=0.32\textwidth]{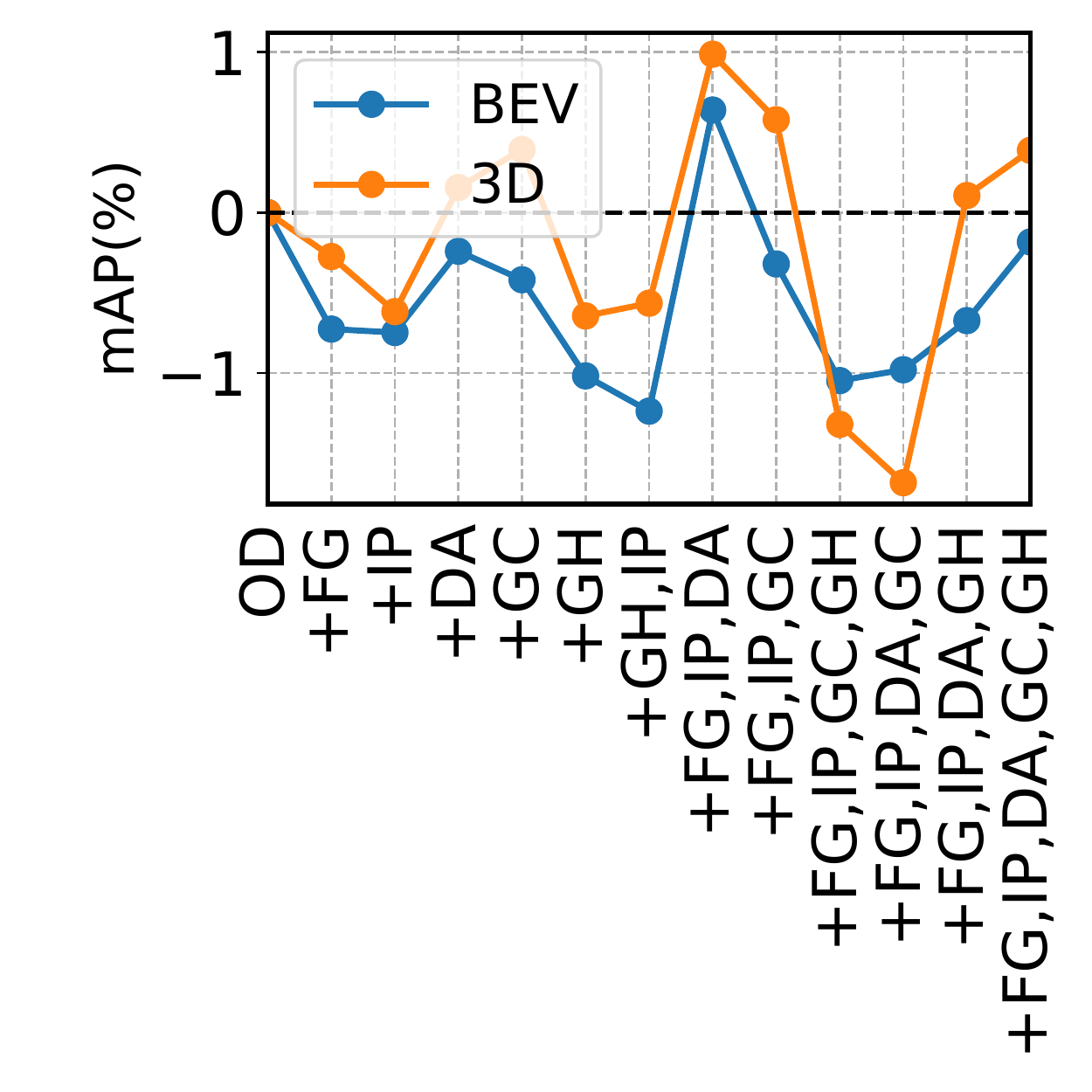}}
    \subfigure[FG + Multi-tasks]{\label{fig:foreground_classification_mtl_ablation}\includegraphics[width=0.32\textwidth]{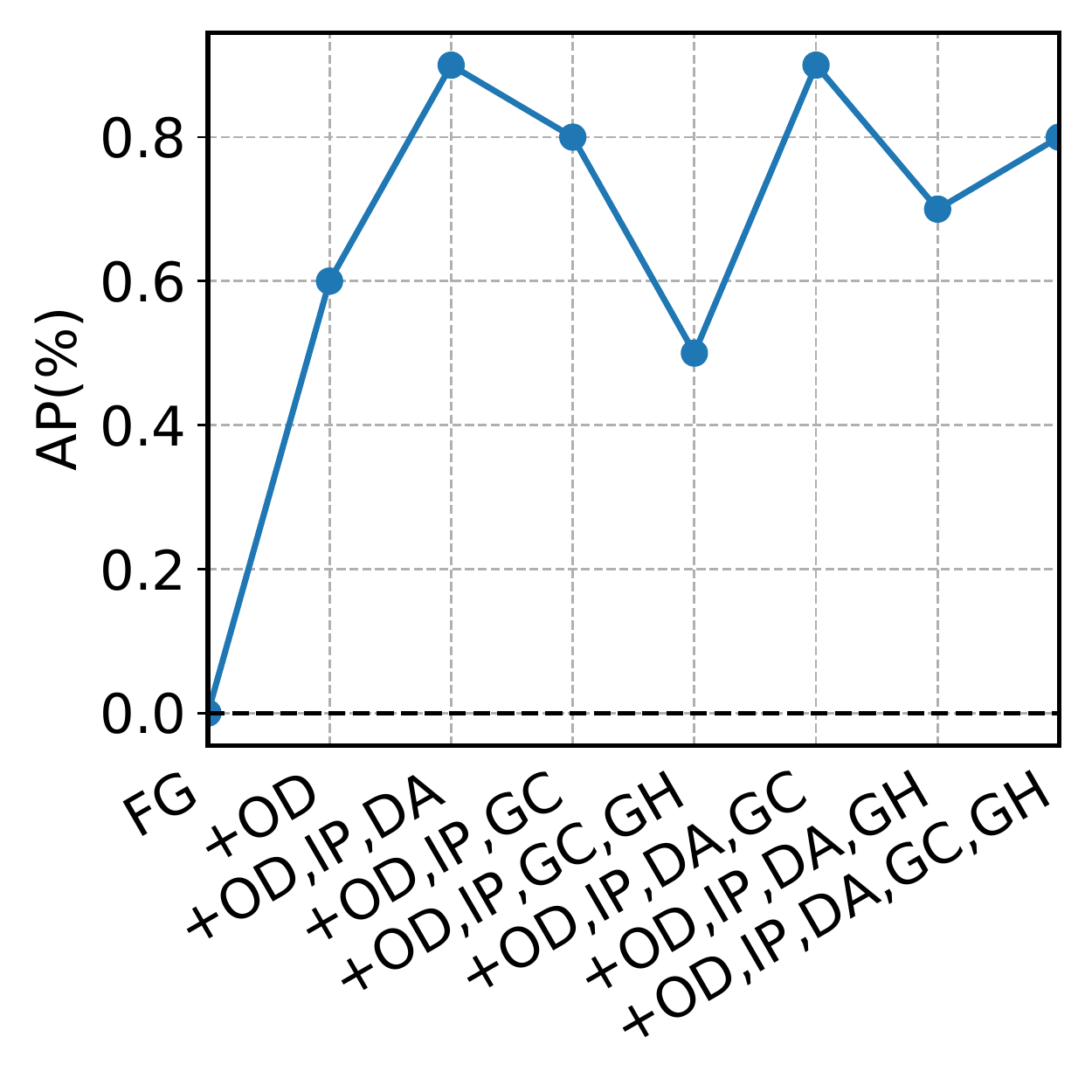}}
    \subfigure[GH + Multi-tasks]{\label{fig:ground_height_mtl_ablation}\includegraphics[width=0.32\textwidth]{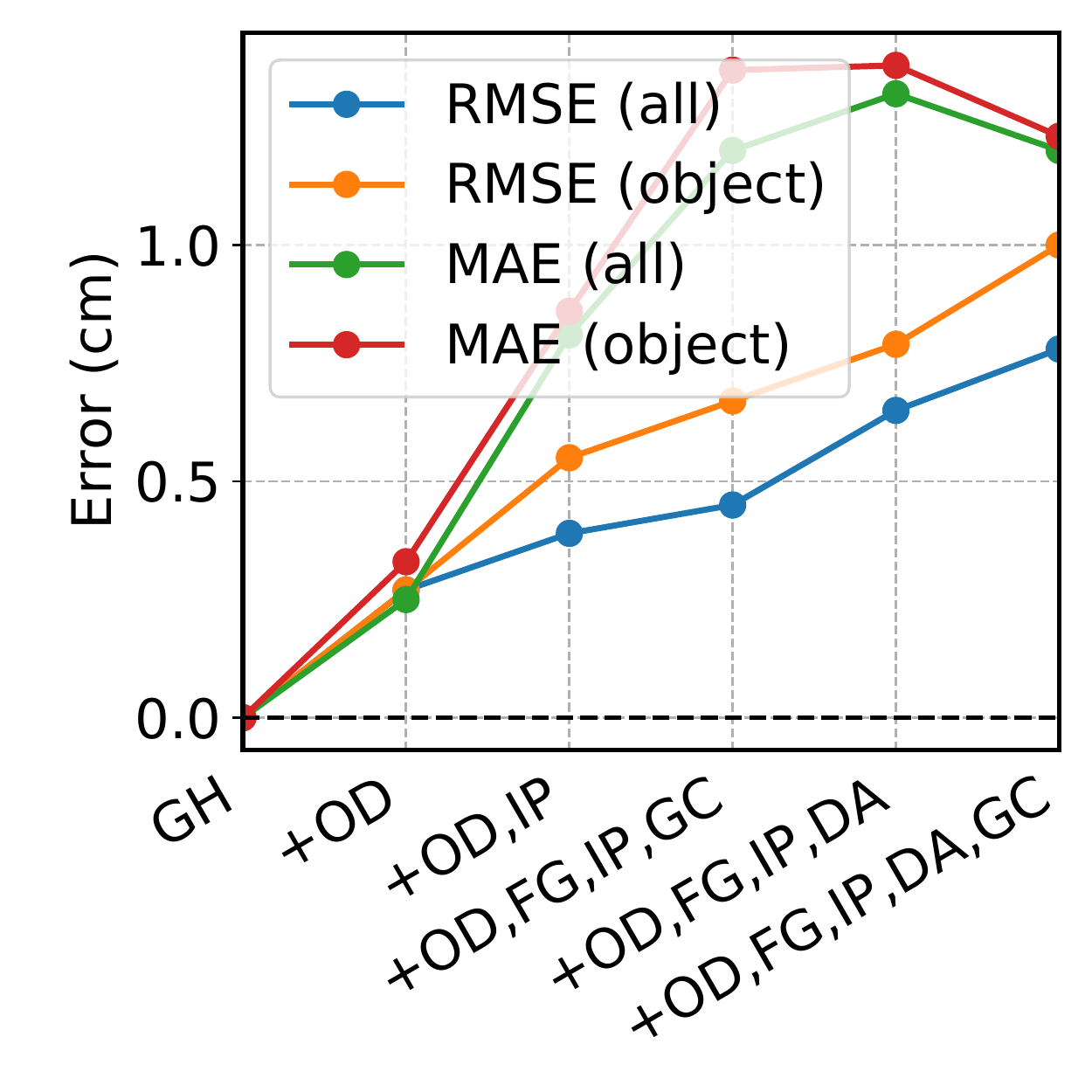}}
	\caption{The performance of the target task from the multi-task network trained with increasing number of auxiliary tasks, relative to the single-task network.} \label{fig:mtl_ablation_study}
\end{minipage}
\vspace{-5mm}
\end{figure}

This section studies the performance of the single task which we focus on (``target task''), with increasing number of multiple tasks (``auxiliary tasks'') in the LidarMTL network. Fig.~\ref{fig:detection_mtl_ablation}, Fig.~\ref{fig:foreground_classification_mtl_ablation}, and Fig.~\ref{fig:ground_height_mtl_ablation} select object detection, foreground classification, and ground height estimation as target task, respectively. We report the perception performance from the multi-task network relative to the singe-task network, with increasing number of auxiliary tasks from left to right on the x-axis. No clear tendency is observed between the object detection performance and the number of tasks. The mAP scores fluctuate between $-1.5\%-1\%$. Introducing more auxiliary tasks increases AP for foreground classification, as well as regression errors for ground height estimation. However, the difference is small (less than $1\%$ AP and $1.5$cm errors). In conclusion, we could achieve on-par single-task perception performance, regardless of the combination of multiple tasks. 

\begin{table*}[tbp]
	\centering
	\resizebox{1\linewidth}{!}{\begin{tabular}{l|cc|c|c|c|cc|cc|c}
			\Xhline{2\arrayrulewidth}
			\multirow{2}{*}{Loss weights} & \multicolumn{2}{c|}{OD} & FG & DA & GC & \multicolumn{2}{c|}{GH} & \multicolumn{2}{c|}{IP} & Avg.\\ \cline{2-10}
			& $\text{mAP}_{BEV} (\%)$ & $\text{mAP}_{3D} (\%)$ & $AP (\%)$ & $AP (\%)$ & $AP (\%)$ & RMSE (cm)& MAE (cm)& RMSE & MAE & Rank \\ \Xhline{2\arrayrulewidth}
			Fixed (equal weights) & $49.6$ & $34.5$ & $96.7$ & $97.7$ & $99.6$ & $20.5$ & $10.7$ & $10.7$ & $6.4$ & $2.7$\\
			Fixed (balanced) & $48.4$ & $32.9$ & $97.0$ & $97.8$ & $99.6$ & $19.2$ & $9.2$ & $12.7$ & $8.1$ & $3.0$\\ 
			Fixed (grid search) & $49.2$ & $34.7$ & $97.2$ & $97.5$ & $99.6$ & $18.6$ & $8.7$ & $10.0$ & $5.7$ & $1.8$\\
			Adaptive~\cite{cipolla2018multi} & $49.2$ & $34.7$ & $97.0$ & $97.2$ & $99.6$ & $24.0$ & $14.1$ & $10.8$ & $6.5$ &$3.3$ \\
			Adaptive~\cite{cipolla2018multi} + grid search & $49.8$ & $35.0$ & $97.0$ & $97.4$ & $99.6$ & $18.6$ & $8.8$ & $9.9$ & $5.6$ & $1.5$ \\			\Xhline{2\arrayrulewidth} \end{tabular}}
	\caption{A comparison among the LidarMTL networks trained with different loss weights.} \label{tab:loss_weights}
\vspace{-5mm}
\end{table*}
\subsubsection{Impact of Loss Weights}
It is known that a proper selection of loss weight for each single task is crucial for multi-task learning~\cite{vandenhende2020multi}. In this ablation study, we train the LidarMTL network with different combinations of loss weights, and compare their multi-task performances. ``Fixed (equal weights)'' assumes that each loss weight is equal. ``Fixed (balanced)'' balances the losses to the similar scales. ``Fixed (grid search)'' finds a set of loss weights by grid search on the training dataset. Note that the loss weights from those three methods are fixed, and do not change during training (cf. Eq.~\ref{eq:standard_mtl_loss}). Instead, ``Adaptive'' employs the uncertainty weighting strategy shown by Eq.~\ref{eq:uncertainty_aware_mtl_loss} to balance single-task losses adaptively. ``Adaptive + grid search'' first puts a set of pre-defined loss weights from the grid search, and then balances the learning with uncertainty weighting.

We report the perception performance for each single task as well as the averaged network's ranking in Tab.~\ref{tab:loss_weights}. Surprisingly, ``Fixed (balanced)'' shows inferior performance even slightly worse than ``Fixed (equal weights) on the averaged ranking, indicating that simply balancing losses might not be the optimal choice in multi-task learning, as different single tasks may have different learning paces. ``Adaptive'' ranks last, with $4-6$cm larger ground height errors compared to the best results, showing the challenge to learn a proper set of loss weights from scratch. The networks trained with loss weights from grid search depict visible improvements (e.g. comparing ``Fixed (grid search)'' with ``Fixed (equal weights)''. When combining uncertainty weighting and grid search, the network slightly outperforms the ``Fixed (grid search)'' strategy, and achieves the best multi-task performance. We conclude the necessity of using loss weights with grid search. 
\subsubsection{Robustness Testing}
\begin{figure}[!tpb]
	\centering
	\includegraphics[width=0.96\linewidth]{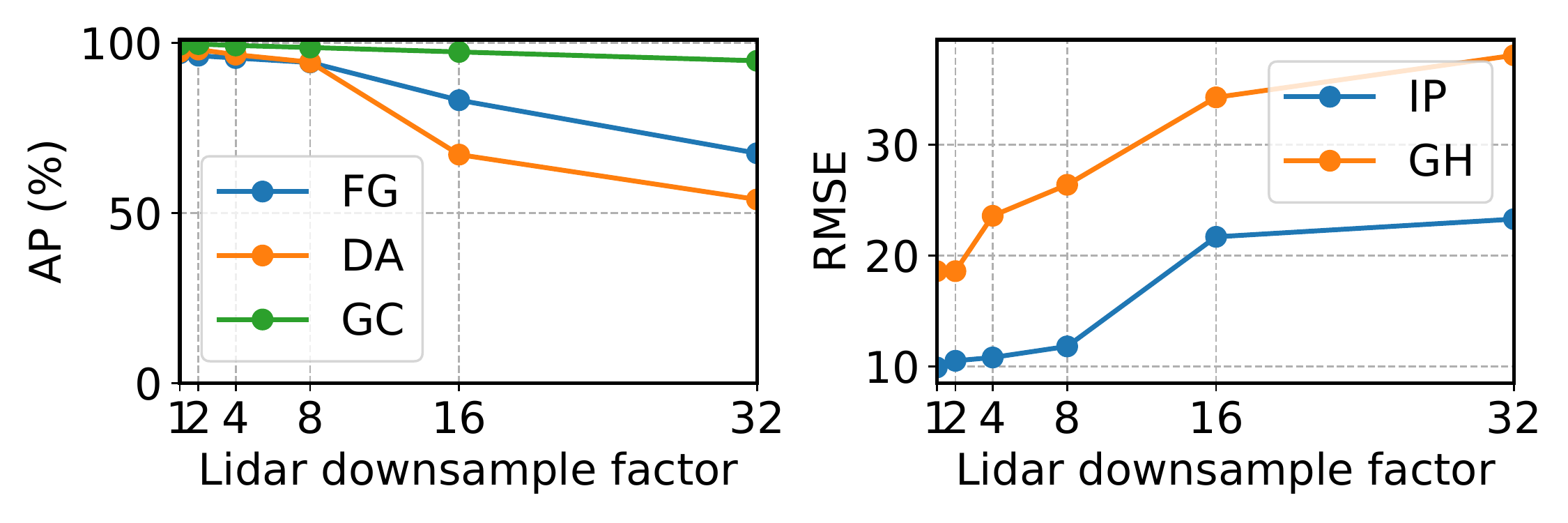}
	\caption{The performance of point-wise predictions (tasks FG, IP, DA, GC, GH) with increasingly sparse Lidar points.} \label{fig:robustness_testing}
	\vspace{-5mm}
\end{figure}
Finally, we study the robustness of point-wise prediction tasks with increasingly sparse Lidar points. Evaluating this robustness is crucial for autonomous driving, because the sparsity of point cloud varies significantly among Lidar sensors and vehicle setup, and largely affects the perception performance~\cite{feng2020labels2}. In this regard, we use the LidarMTL network trained with full Lidar points (8000 voxels) to make inferences on the evaluation data with downsampled Lidar points by factors $2,4,8,16,32$. Results are shown in Fig.~\ref{fig:robustness_testing}. The performance of DA, GC and IP drops slightly at downsample factors smaller than $8$. FG remains high AP scores above $90\%$ even at $32$ downsample factor (i.e. $250$ non-empty voxels). The performance of GH drops quickly at downsample factor $4$. The experiment shows that different tasks have different robustness against Lidar point cloud sparsity.
\subsection{Application to Online Localization}\label{experimental_result:application_online_localization}
Localization in urban environments requires point cloud maps and point registration algorithms. However, Lidar-based localization typically suffered from dynamic objects and undulating road surfaces\cite{urbanloco}. By semantically segmenting the scene, LidarMTL provides an ideal pre-processing for such localization modules. To study LidarMTL's impact on localization, we used the outputs from DA and FG to help localizing the vehicle. As a comparison, we have 4 types of inputs for the localization algorithm: the raw scan, the point cloud without DA, the point cloud without FG, and the point cloud without both DA and FG. 

We perform localization experiments on 24 trajectories spanning 2.69KM, and the vanilla NDT registration algorithm in Autoware \cite{autoware1} was chosen as the real-time localization module. In our experiment, the NDT voxel resolution is 1 meter. The map is created with the ground-truth scan without DA and FG downsampled to 0.2 meters. 

Tab.~\ref{tab:online_localization_results} shows the performance of the localization algorithm with various point cloud input. We also included the result from ground-truth DA and FG tasks for a comparison. The performance is evaluated with Root Mean Squared Error along three axis and the yaw angle. Furthermore, we listed the success rate for these tests: one is considered as a failure if the translation RMSE is larger than 3m or if the rotation RMSE is larger than 4$^{\circ}$. Compared with the raw input, the LidarMTL-processed inputs yield more accurate localization. Furthermore, since dynamic objects are removed from the scan, the algorithm performed robustly in complicated environments. As compared with the ground truth point cloud inputs, the LidarMTL pre-processing reaches similar level of localization accuracy and success rate. Given the stochastic nature of the NDT algorithm, the LidarMTL even out-performed ground truth segmentation in certain evaluation matrices.

\begin{table}[tbp]
\centering
\resizebox{1\linewidth}{!}{

\begin{tabular}[t]{l|l|ccc|c|c}
\Xhline{2\arrayrulewidth}

\multicolumn{2}{c|}{Point Cloud} & \multicolumn{3}{c|}{Translation RMSE (m)} & Rotation RMSE ($^{\circ}$) & Success\\ \cline{3-6}
\multicolumn{2}{c|}{Type}& X & Y & Z & Yaw & Rate ($\%$)\\
 \Xhline{2\arrayrulewidth}  
\multicolumn{2}{c|}{Raw} & 1.80 & 1.27 & 1.13 & 1.16 & 83.3\\
\cline{1-6}
& no DA& 1.13 & 0.67 & 0.34 & 1.00 & 95.8 \\
GT& no FG& 1.46 & 0.56 & 0.46 & 0.71 & 95.8\\
& no DA, FG& 1.51 & 0.87 & 0.34 & 1.25 & 95.8 \\
\cline{1-6}
&no DA& 1.83 & 0.59 & 0.21 & 0.84 & 95.8\\
LidarMTL &no FG& 1.62 & 0.58 & 0.43 & 0.82 & 95.8 \\
&no DA, FG& 1.73 & 0.65 & 0.06 & 1.13 & 100\\
\Xhline{2\arrayrulewidth} 
\end{tabular}}
\caption{Online localization results.}
\label{tab:online_localization_results}
\vspace{-5mm}
\end{table}
\section{Discussion and Conclusion} \label{sec:conclusion}
We have presented the multi-task network to jointly performs six perception tasks for 3D object detection and road understanding, which were only studied separately in previous works and lack quantitative analysis. Comprehensive experiments verified the network's design and the multi-task performance. The proposed multi-task network is small, fast, accurate, and useful for localization, making it highly desirable for online deployment in autonomous cars. In the future, we plan to extend the network to process multiple frames for motion estimation.


\bibliographystyle{IEEEtran}
\bibliography{bibliography}

\end{document}